\documentclass[runningheads]{llncs}
\usepackage{graphicx}
\usepackage{amsmath}
\usepackage{amssymb}
\usepackage{booktabs}
\usepackage{wrapfig}
\usepackage{cite}
\usepackage{soul}
\usepackage{xcolor}
\usepackage{pifont}
\usepackage{multirow}
\usepackage{algorithm2e}
\usepackage{nicefrac}
\usepackage{comment}
\RestyleAlgo{ruled} 
\usepackage{caption}
\usepackage{subcaption}
\usepackage{microtype}
\DeclareMathOperator*{\argmax}{arg\,max}

\begin{document}
\title{Weakly supervised information extraction from inscrutable handwritten document images}
%
%
\author{Sujoy Paul \and Gagan Madan \and Akankshya Mishra \and Narayan Hegde \and Pradeep Kumar \and Gaurav Aggarwal}
\authorrunning{S. Paul et al.}
\institute{Google Research}

\titlerunning{Weakly supervised information extraction}
%
%
\maketitle              
\begin{abstract}
State-of-the-art information extraction methods are limited by OCR errors. They work well for printed text in form-like documents, but unstructured, handwritten documents still remain a challenge. Adapting existing models to domain-specific training data is quite expensive, because of two factors, 1) limited availability of the domain-specific documents (such as handwritten prescriptions, lab notes, etc.), and 2) annotations become even more challenging as one needs domain-specific knowledge to decode inscrutable handwritten document images. In this work, we focus on the complex problem of extracting medicine names from handwritten prescriptions using only weakly labeled data. The data consists of images along with the list of medicine names in it, but not their location in the image. We solve the problem by first identifying the regions of interest, i.e., medicine lines from just weak labels and then injecting a domain-specific medicine language model learned using only synthetically generated data. Compared to off-the-shelf state-of-the-art methods, our approach performs $>2.5\times$ better in medicine names extraction from prescriptions. 


\end{abstract}
\section{Introduction} \label{sec:intro}

\begin{figure}
    \centering
    \includegraphics[scale=0.35]{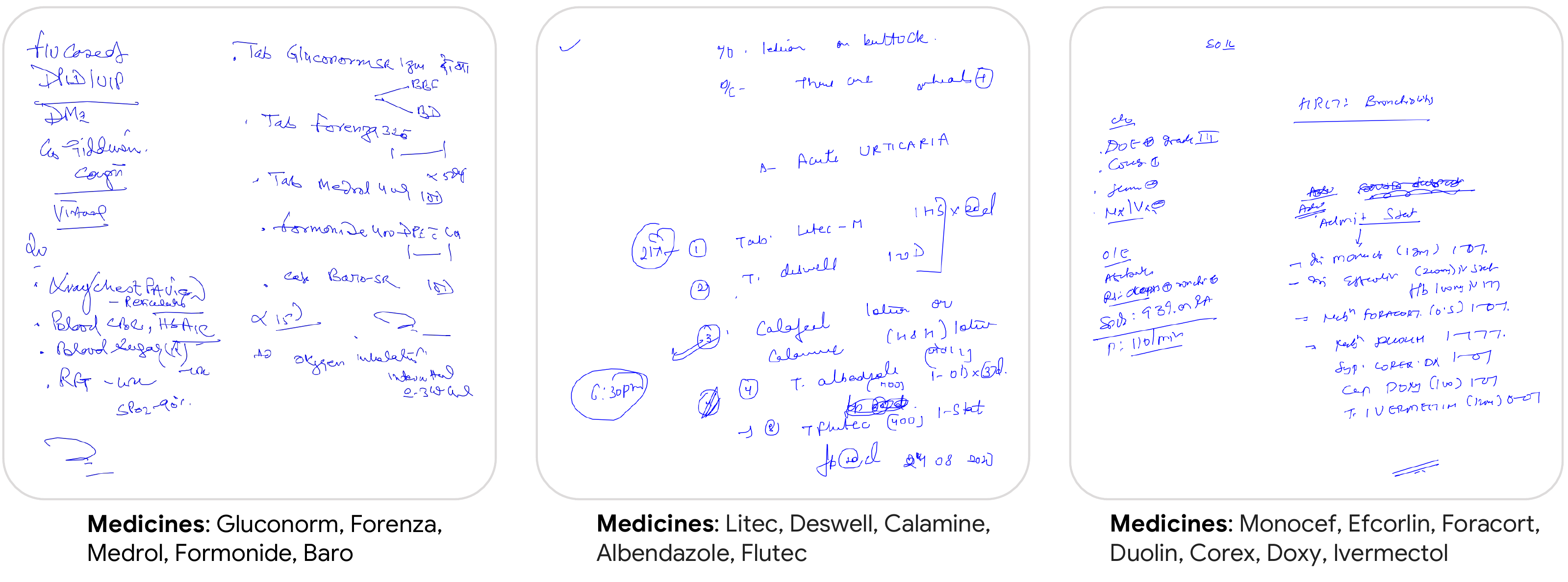}
    \caption{Samples representative images from the prescription dataset used in this work. As we can see the handwriting is often inscrutable and does not follow any specific structure or format. The task we focus in this paper is to extract medicine names from such images.}
    \label{fig:sample_images}
\end{figure}

Optical character recognition (OCR) enables the translation of any image containing text into analyzable, editable and searchable format. Over the last decade, many large scale models \cite{li2021trocr,diaz2021rethinking,ingle2019scalable} and sophisticated techniques \cite{breuel2013high, bissacco2013photoocr, long2021scene} have been developed with neural network based architectures for OCR. These systems are not only limited to printed text but also work quite well on handwritten text, as they are trained on large amount of labeled as well as synthetic handwritten data. In the past, there have also been works around developing domain specific OCR models \cite{bukhari2017anyocr, thompson2015customised, karthikeyan2021ocr}. Most of these works develop these models for generic text lines \cite{marti2002iam, karatzas2015icdar}, and require meticulously labeled data for learning. In this work, we primarily focus on how we can improve the quality of existing OCR models on very hard to read, unstructured documents for specific entities of interest, with an application in handwritten medical prescriptions. 

In many countries, prescriptions are primarily delivered to patients in handwritten formats by doctors. A few billion prescriptions are generated every year world-wide \cite{jayakumar2021online}. Digitizing them would unlock numerous applications for many stakeholders and use cases in the healthcare ecosystem like e-pharmacies, insurance companies, creating electronic health records necessary for preventive healthcare, better diagnosis, analysis at a local and global level for policy making and so on. However, most of such documents, as shown in Figure \ref{fig:sample_images} are often hard to read for non-pharmacists \cite{pragnadyuti2017legibility}. Even pharmacists go through months/years of training to decipher such prescriptions. Existing state-of-the-art OCR models though trained on large amount of data, do not perform well on such inscrutable documents. Procuring large domain-specific datasets is not a cost-effective or scalable solution, as it involves annotation that too from domain experts which can become quite expensive. Although there have been some works \cite{gupta2021algorithms,achkar2019medical,rani2022recognition} in extracting information from handwritten prescriptions, the algorithms are not generalizable, heavily hand-tuned and lack rigorous evaluations. 
With these problems in mind, we propose an approach that can significantly enhance the performance of existing state-of-the-art OCR systems by selectively infusing domain knowledge using only weak supervision. 

Medical prescriptions consist of various information like data from lab reports, ordered tests, health vitals, observations along with medicine names. Our work focuses on the medicine section which is considered the most important from a consumer standpoint, but the techniques can be similarly applied to other sections or other types of documents beyond prescriptions, such as printed forms filled with handwriting. The medicine section of a prescription has a rough semantics consisting of medicine name, category, frequency of intakes and quantity (see Figure \ref{fig:sample_images}). As these are non-form type of documents and quite unstructured, it is a challenge to extract medicine name entities from such documents.

Most OCR approaches \cite{ingle2019scalable, li2021trocr} take a two step approach - first localize the text regions by detecting bounding boxes around them, and then recognizing the text using line recognition models. The recognition model often consists of an optical recognizer and a language model (LM) to correct the optical model errors. The LM gives us the flexibility to infuse domain-specific knowledge. But, injecting such knowledge to all lines in the document may not be optimal, as different parts of the document can correspond to different entities, or even domains. For example, the pattern in which a medicine name is written is very different from the pattern in which normal text such as observations are written in the same prescription. Thus, in order to enhance the recognition of medicine names and extract them from the prescription, we first detect lines where medicine names are written. Then in the recognition model, we inject a LM which is specific to medicine names. For the rest of the image, we inject the vanilla LM.

Note that to learn the model which detects medicine lines, we do not use strong bounding polygon labels, but rather only weak labels, i.e., the medicine names present in the image. Such weak labels are much easier to obtain, as the annotators do not need to draw a bounding polygon and often labeling comes for free, for example, when a medicine bill is paired with a prescription. Apart from that, to learn the medicine LM, we do not use any annotated text lines, but rather generate synthetic text lines using a probabilistic programming approach. Our weakly supervised medicine line detector obtains 78\% pixel mIoU with just weak labels, and helps to selectively infuse medicine LM, which in turn improves the overall performance from 19\% to 48\% jaccard index. The main contributions of this work are:
\begin{itemize}
\item Develop a weakly supervised segmentation method to detect specific text entities, such as medicine names in handwritten prescriptions.
\item Learning a domain-specific medicine LM using synthetic medicine name lines generated by probabilistic programs and using it to enhance the performance of state-of-the-art OCR models. 
\item A model dependent unique way of enhancing the performance of matching with words from the vocabulary. 
\end{itemize}

\section{Related Works} \label{sec:rel}

\subsubsection{Optical Character Recognition}
OCR literature has seen tremendous improvements in the past decade. The successes \cite{li2021trocr,diaz2021rethinking,ingle2019scalable} can be attributed to sophisticated models, synthetic data generation, various augmentation techniques, among others. An OCR system is made of multiple models, starting from text detection \cite{long2021scene,wang2021towards,long2022towards}, script identification \cite{fujii2017sequence,huang2021multiplexed}, and finally line recognition \cite{bhunia2021text,li2021trocr,diaz2021rethinking,litman2020scatter}. Even with all these advancements, recognition of handwritten lines still remains a challenging task as writing style can be a unique signature of the person, allowing room for huge variations. In our experiments, we found that off-the-shelf line recognition models, even though perform quite well for a lot of printed and handwritten datasets, they fail to perform equally well on handwritten images. In this work, we show how we can improve their accuracy by more than 2 times the baseline by first detecting specific entities of interest (rather than detecting all text) and then improving the line recognition model by injecting domain-specific LMs. We next discuss the existing literature around these topics. 

\subsubsection{Weakly-supervised Detection}
Detecting specific entities of interest in an image can be posed as detection or segmentation task. However, to learn these tasks, traditional methods would need strong labels, i.e., either pixel-wise \cite{cheng2020panoptic,long2022towards} or bounding box labels \cite{ren2015faster,redmon2016you,long2022towards}. In the recent past, there has been a lot of work in developing methods which can learn from only weak labels, such as weakly-supervised object detection \cite{zhang2021weakly,li2016weakly}, segmentation \cite{khoreva2017simple,wei2016stc}, action detection \cite{paul2018w,zhang2021cola}, etc. These methods do not need access to strong labels such as bounding boxes, but can learn from just weak labels, i.e., image-level labels of the object categories present in the individual training images. Such a formulation reduces the manual labor needed to acquire strong labels, thus making it scalable to large datasets. 

Motivated by these, we aim to learn a segmentation model to detect entities of interest in an image, such as medicine names from just weak labels, i.e., list of medicine names given an image. In this use case, the individual entities do not correspond to any underlying category unlike segmentation or detection of objects in natural scenes. Recently, it has been shown \cite{kittenplon2022towards} that using weakly labeled data along with strong labels improves the performance of scene text recognition. In our task, we only have weakly labeled data without any strong labels (synthetic or real) and the text is primarily handwritten which is often inscrutable even if text detection is perfectly done. Moreover in our use case, we need to detect specific entities among other cluttered text, and not any generic text. There are also works on defining rules to derive weak labels from the data \cite{ratner2017snorkel}. While that is quite challenging and not generalizable in our use case, we use the intuition to convert the weak labels to strong labels via labeling functions.

\subsubsection{Domain-specific Language Models}
There has been a lot of work \cite{lee2019biobert,rasmy2021med,yang2022gatortron} which shows that injecting domain-specific knowledge in LMs helps to perform much better on those domains than models developed on generic text. Specifically for OCR, there have been some works \cite{d2017generating, gupta2021unsupervised} showing that having access to domain related text data helps to adapt existing LMs and thus improving final OCR performance. However, in our use case of decoding medicine names, it is non-trivial to acquire lines of medicine names written by doctors, as they are hardly available in normal text corpus. To solve that, we use domain knowledge to define a probabilistic program which can take in the medicine name and generates patterns of medicine lines as would be written by doctors in prescriptions. We show that using such a LM in the OCR decoder improves the performance significantly.

\section{Methodology} \label{sec:method}

\begin{figure}
    \centering
    \includegraphics[scale=0.3]{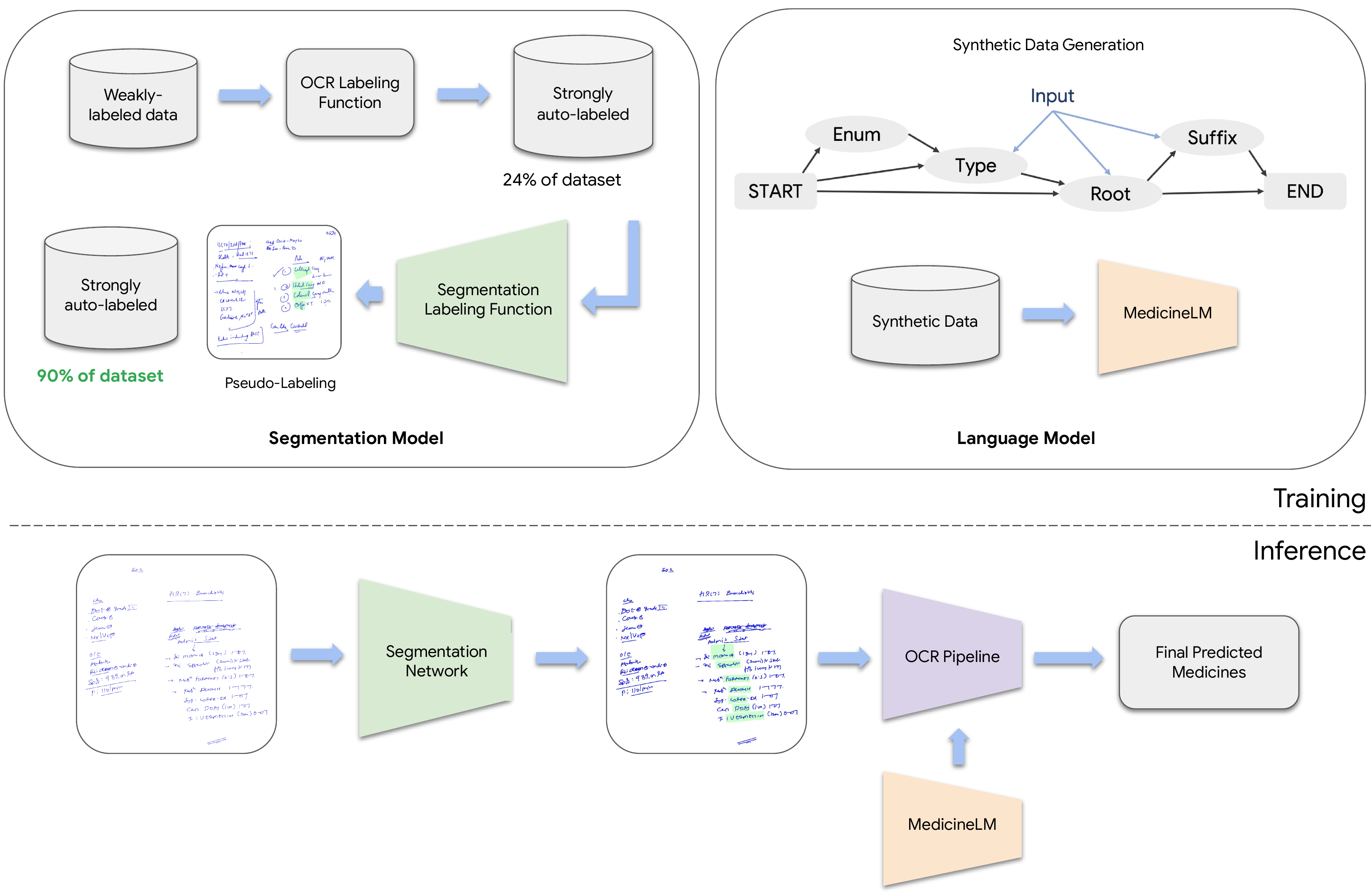}
    \caption{\small Training and inference pipelines for medicine name extraction from prescriptions. The top-left block shows the weakly supervised medicine line segmentation pipeline. The top-right block shows the process of generating synthetic medicine lines using probabilistic programs and then using it to train a medicine LM. The bottom row shows the inference pipeline, that first localizes the medicine names using the segmentation network, and then injects the medicine LM while decoding the OCR outputs.}
    \label{fig:framework}
\end{figure}

\subsection{Problem Statement}
In this work, we focus on the problem statement of extracting textual entities from non-form type handwritten document images, which are often hard to read. We specifically focus on the challenging problem of extracting medicine names from handwritten prescriptions as shown in Figure \ref{fig:sample_images}. Formally, given an image $\boldsymbol{x}$, the output of the framework should be the medicine names $\{m_j\}_{j=1}^n$ that appear in the image, where $m_j \in \mathrm{V}$, the vocabulary of medicines. $n$ denotes the number of medicines in the prescription that varies from prescription to prescription.  The training data that we use to solve this problem is only weakly labeled, i.e. for every image, we have a list of medicine names that appear in the image, and not their bounding box locations. Thus, our training data contains tuples of image and unordered set of medicine names as follows, $\mathcal{D} = \{(\boldsymbol{x}_i, \mathcal{G}_i=\{m_j\}_{j=1}^{n_i})\}_{i=1}^N$, where $n_i$ denotes the number of medicines in that image, $N$ denotes the number of images in the training data and $\mathcal{G}_i$ is the ground truth list of medicines. 

\subsection{OCR Line Recognition Model}
Most line recognition models have two parts - the encoder, often called the optical part of the model, which encodes the visual information, and the decoder, which is either trained end-to-end with the encoder, or CTC type decoder \cite{graves2006connectionist} where the encoder outputs are combined with LM scores to obtain the final text. We use the second option and train our network with CTC loss \cite{graves2006connectionist}. This allows us to decouple the optical and the LM, and replace it with domain specific LMs.

\subsubsection{Encoder:}
The encoder or the optical part of the line recognizer consists of first 7 layers of inverted bottleneck conv layers \cite{sandler2018mobilenetv2} with 64 filters and stride of 1, followed by 12 layers of transformer encoder \cite{vaswani2017attention} with hidden size of 256 and 4 attention heads, and finally a fully connected symbol classification head. We use this backbone from \cite{diaz2021rethinking}, as it achieves state-of-the-art performance on various datasets. Our pre-trained model is also the same as \cite{diaz2021rethinking}. It is interesting to note that our method is agnostic to the encoder used as it can be used to boost the performance of any OCR backbone. 

\subsubsection{Decoder:}
We use a CTC decoder \cite{graves2006connectionist} following \cite{diaz2021rethinking}, which combines scores from the encoder logits and a character n-gram LM. We set $n=9$ unless otherwise mentioned. We will discuss how we train and use a medicine LM subsequently.

\subsection{Weakly Supervised Line Segmentation}
We next discuss our algorithm to detect medicine lines by just using weak labels while training, i.e., only the medicine names for every image, and not their bounding polygons. Note that while we use this method for medicine line detection, it can be also used for detecting other entities in other document types.

\subsubsection{Labeling Functions} 
At the core of our algorithm is the idea of using labeling functions to automatically convert a weakly labeled dataset to strongly labeled. There have been some works \cite{ratner2017snorkel} in literature where rules are defined as labeling functions. The labeling functions may not be as perfect as a human oracle and the strong labels they generate may have errors in them. There are often thresholds or rules used to reduce errors. Thus, while defining a labeling function we need to optimize coverage, which is the number of data points that can be labeled using such labeling functions and their error rate. Although there can be some noise in such labeling, this significantly reduces the annotation cost. We sequentially apply two labeling functions, as discussed next to convert a list of medicine names to bounding boxes. In our use case of assigning a bounding box to each medicine name, we can consider it as an assignment problem between the detected bounding boxes ($p$) by a generic text detector and the number of medicines in it $n$. Considering $p=50$ and $n=5$, the number of possible assignments turns out to be ${}^pC_n {}^pP_p \approx 2.5e8$. We solve this problem via two techniques - using the content of the boxes (via OCR Labeling Function), and using the visual features (via Segmentation Labeling Function). 

\subsubsection{OCR Labeling Function:}
As for every image, we have the list of medicines that appear on it, for each detected word in the image, we can naively find the closest medicine name (by edit distance) from the ground truth list, albeit applying a threshold. However, directly using the edit distance may not respect the model's predictions. For example, according to the OCR line recognition model, modifying an $i$ to $l$ may have lower cost than $i$ to $z$, but it would be the same edit distance for both the cases. Thus, in order to utilize the model's predictions, we decode up to the top-k predictions, and stop when we find an exact match with a medicine name from the list of ground truth medicines, i.e., the weak labels. The bounding box associated with these matched words then can be used as the ground-truth bounding boxes of medicine names. We can define the labeling function as $\mathcal{F}(\boldsymbol{x}) = \{(t_j, l_j, h_j, w_j, r_j)\}_{j=1}^{q}$, where the bounding boxes of $m$ medicines are in the rotated box format and $t_j, l_j, h_j, w_j, r_j$ representing top, left, height, width, and rotation angle of each matched bounding box. Then, we can construct a training dataset as follows: $\mathcal{D}_{tr} = \{(\boldsymbol{x}_i, \mathcal{F}(\boldsymbol{x}_i))\}_{i=1}^N$.

The number of matching bounding boxes $q_i \leq n_i$, as in most cases the handwriting is so illegible that to decipher that even a higher number of top-k lines may not allow a match with the ground truth medicine names. This can happen for a sizable number of images, which in turn can introduce a significant noise in the data, leading to problems in learning the segmentation network. Thus, we only use those images to train our network where we find that at least 90\% of the ground truth medicines have been matched. The reason behind setting such a high threshold is this set becomes the guiding signal for the rest of the algorithm. Thus our modified strongly-labeled training dataset can be represented as: $\mathcal{D}_{tr} = \{\big(\boldsymbol{x}_i, \mathcal{F}(\boldsymbol{x}_i)\big)\Big| \frac{|\mathcal{F}(\boldsymbol{x}_i)|}{|\mathcal{G}_i|} \geq 0.9\}_{i=1}^N$. While increasing the number of top-k paths helps more images to pass this threshold, we find that it saturates after a point, specially for documents which are hard to read, such as prescriptions used in this work. 
While the 0.9 threshold allows us to reduce missing bounding boxes in the training set, it also reduces the number of images in the training set, as $|\mathcal{D}_{tr}| \leq |\mathcal{D}|$. We next discuss a second labeling function to alleviate this problem.

\subsubsection{Segmentation Labeling Function} It may happen that even after decoding a high number of paths ($k$), we still are not able to match all the ground truth medicine names. This can happen when the handwriting is quite challenging for the model to predict accurately. In such a scenario, we leverage the visual appearance features via the segmentation model itself, rather than just labeling via OCR. Motivated by the success of self-training in domain adaptation \cite{liu2021cycle,araslanov2021self} and semi-supervised \cite{sohn2020fixmatch,cascante2021curriculum}, we use the segmentation model to pseudo-label the images in the rest of the dataset, i.e., $\mathcal{D}$ - $\mathcal{D}_{tr}$.

First, we train a segmentation network $\mathcal{M}$ using the relatively small training data $\mathcal{D}_{tr}$ obtained from the OCR Labeling Function outlined above. Then, we use it to predict the medicine lines on the images in $\mathcal{D}$ - $\mathcal{D}_{tr}$. We can consider the output of the model to be $\mathcal{M}(\boldsymbol{x}) = \{(t_j, l_j, h_j, w_j, r_j)\}_{j=1}^{l}$. Following our previous threshold, we add those images to the training dataset, where the union of the number of predicted medicine lines by the segmentation network and the OCR labeling function above, is at least 90\% of total number of medicines in that image. We can represent the new training set as follows: 
$\mathcal{D}_{tr} = \{\big(\boldsymbol{x}_i, \mathcal{F}(\boldsymbol{x}_i) \cup \mathcal{M}(\boldsymbol{x}_i)\big)\Big| \frac{|\mathcal{F}(\boldsymbol{x}_i) \cup \mathcal{M}(\boldsymbol{x}_i)|}{|\mathcal{G}_i|} \geq 0.9\}_{i=1}^N$. 

Ideally, we can repeat this process, i.e. repeat pseudo-labeling the training images using a trained segmentation model and training a new model with the pseudo-labeled training set. The training set would grow over iterations. The two labeling functions can be generalized as: 
$\mathcal{D}_{tr}^T = \{\big(\boldsymbol{x}_i, \cup_{t=1}^T \mathcal{M}_t(\boldsymbol{x}_i)\big)\Big| \frac{\cup_{t=1}^T \mathcal{M}_t(\boldsymbol{x}_i)}{|\mathcal{G}_i|} \geq 0.9\}_{i=1}^N$, where $\mathcal{M}_t = \mathcal{F}$ for $t=1$, and the $t^{th}$ medicine line segmentation model for $t\geq 1$, and $T$ represents the total number of iterations. 
\begin{figure*}[t]
     \centering
     \begin{subfigure}[t]{0.32\textwidth}
        \centering
        \includegraphics[scale=0.09]{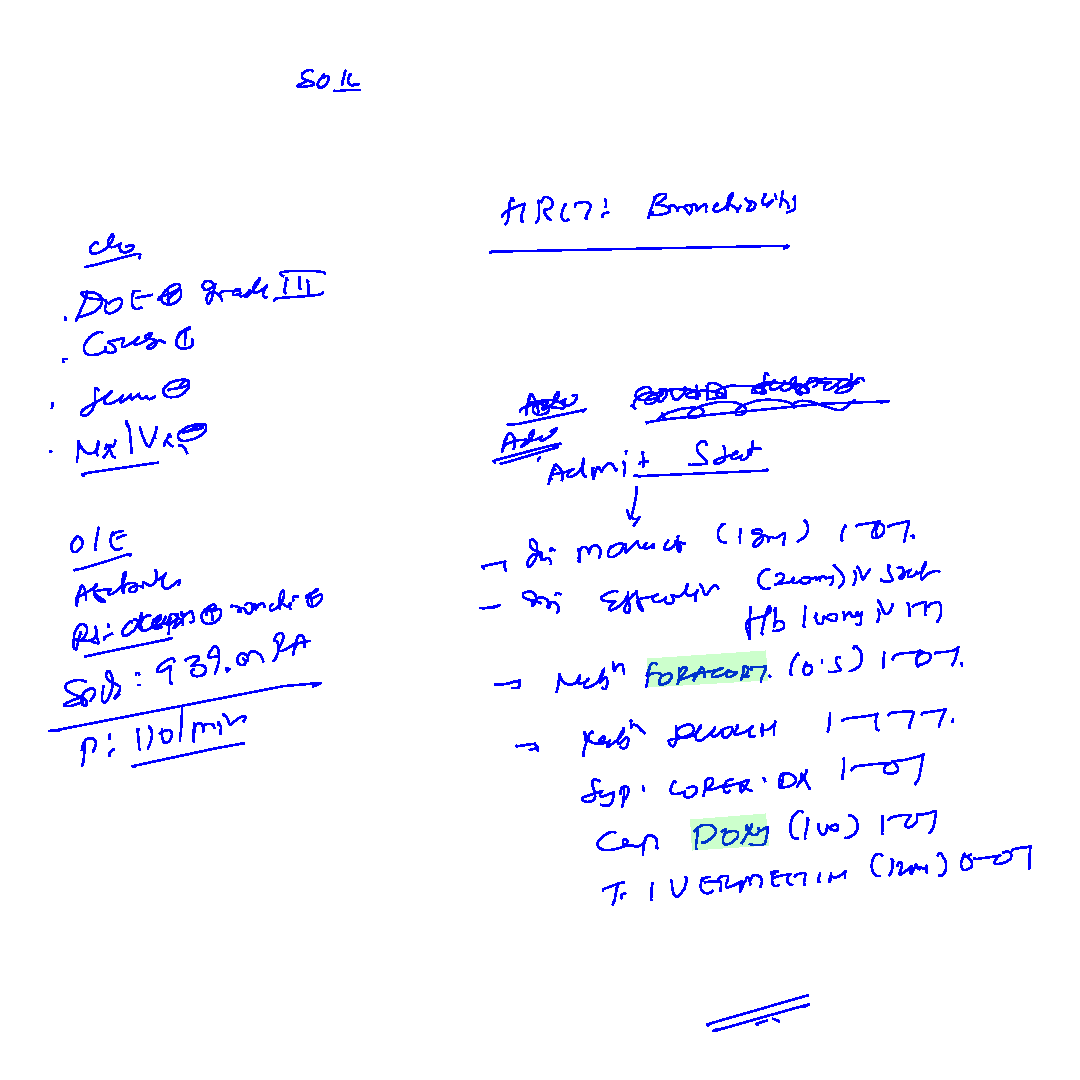}
        \caption{Iter 1}
        \label{fig:anno1}
     \end{subfigure}
     \begin{subfigure}[t]{0.32\textwidth}
         \centering
         \includegraphics[scale=0.09]{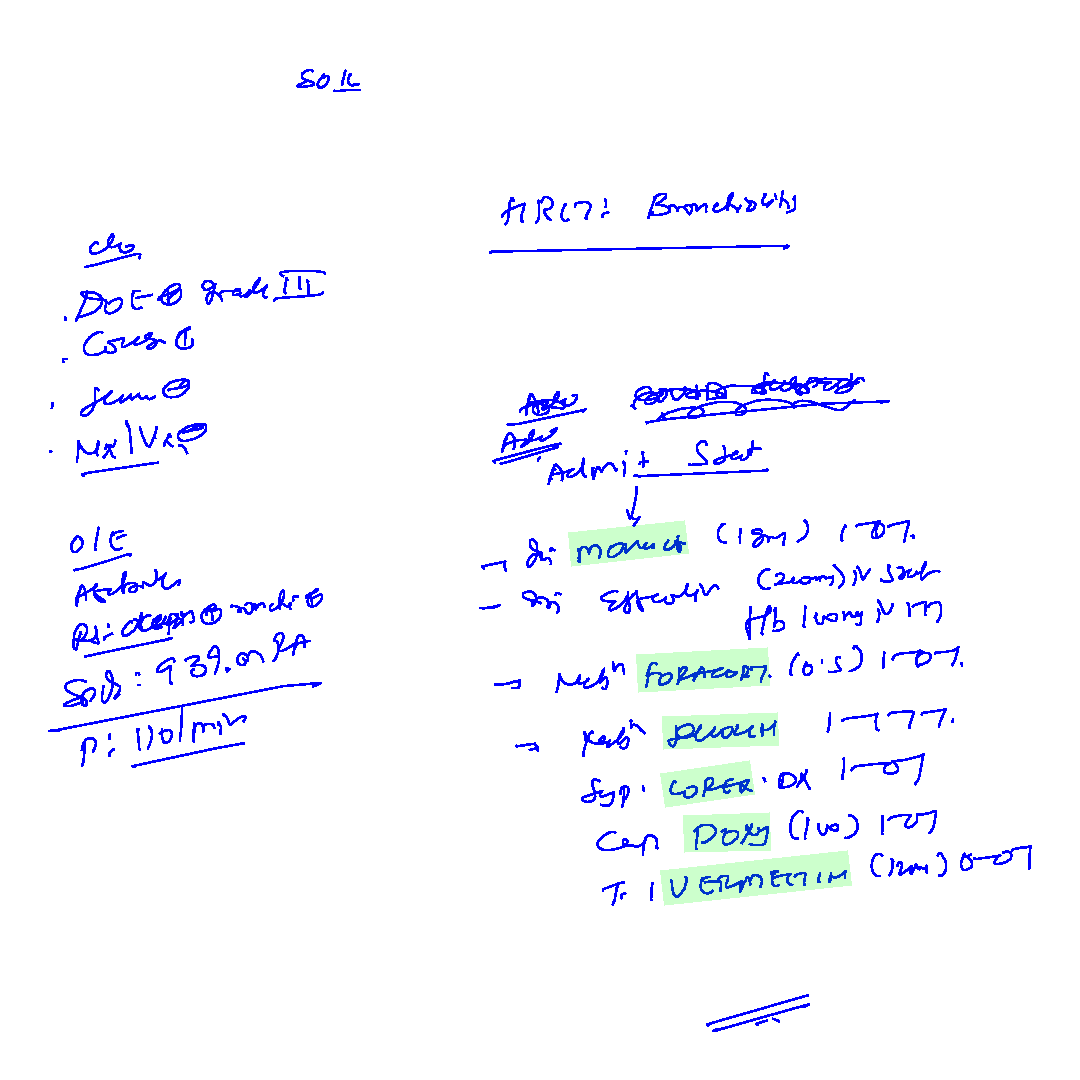}
         \caption{Iter 2}
         \label{fig:anno2}
     \end{subfigure}
     \begin{subfigure}[t]{0.32\textwidth}
         \centering
         \includegraphics[scale=0.09]{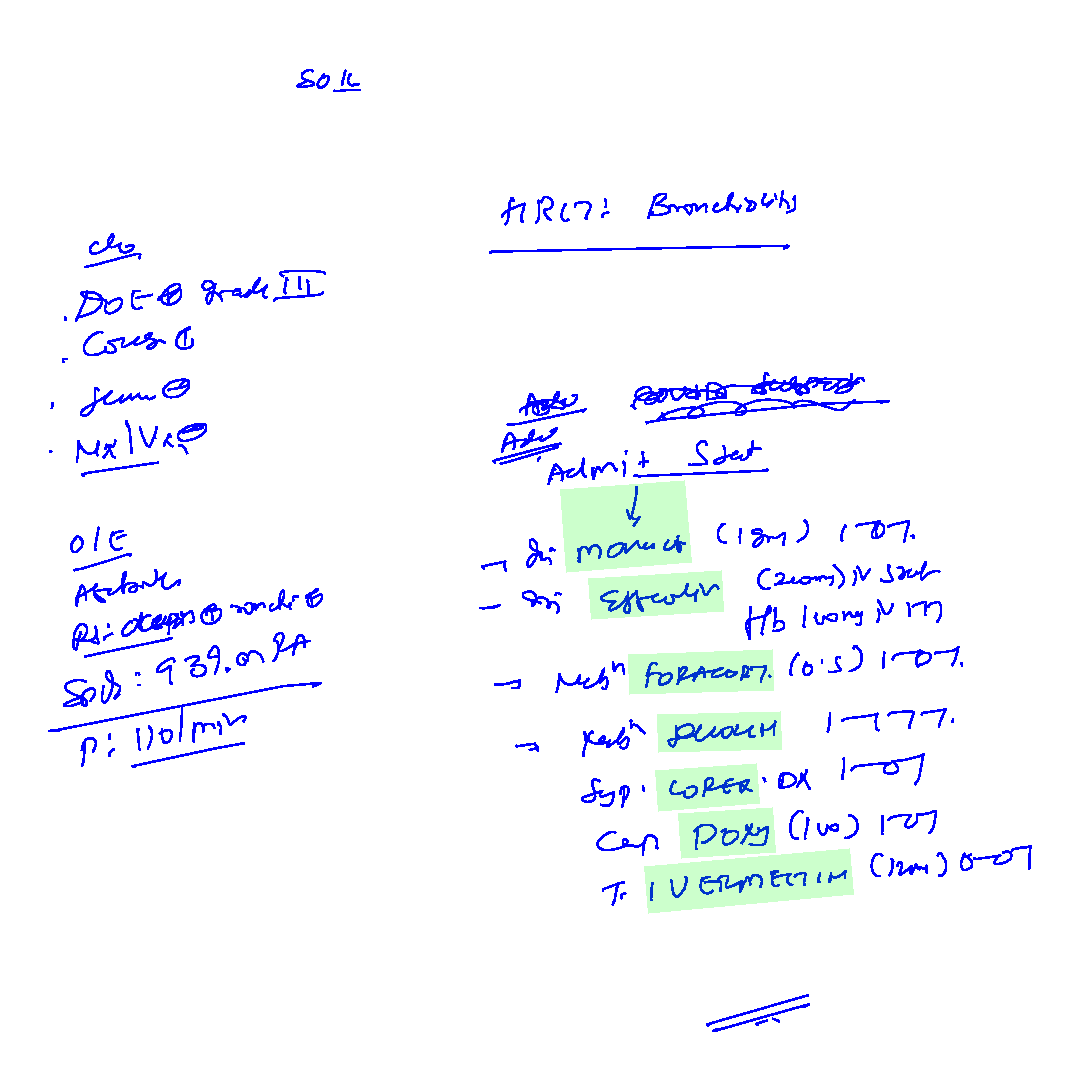}
         \caption{Iter 3}
         \label{fig:anno3}
     \end{subfigure}
     \caption{\small Evolution of labels from the labeling functions. Iter 1 represent the OCR Labeling Function and the subsequent ones represent the Segmentation Labeling Function for different iterations. The green highlighted regions denote the detected medicine names.}
    \label{fig:annotation_quality}
\end{figure*}

Figure \ref{fig:annotation_quality} shows how segmentation improves over iterations. Using only the OCR Labeling Function misses out some of the medicine names, as it is dependent on the ability of the underlying OCR model we use to decipher the medicine names. However, applying the Segmentation Labeling Function on top of it helps to predict the medicine patches which were missed, as it does not depend on OCR or the content, but rather on the visual features, such as strokes, indentation, etc. which we will discuss later in Section \ref{sec:exp}. 

\subsubsection{Segmentation Model}
Given the bounding boxes obtained using the labeling functions, we can train a medicine line segmentation model. Our segmentation model is DeepLab \cite{cheng2020panoptic} with a ResNet50 backbone \cite{he2016deep}. Although we use this architecture, it can be replaced by any other state-of-the-art segmentation model. We convert the bounding boxes to label masks, and use them as supervision to train the segmentation network. The label mask has either 0 or 1 at each pixel location, denoting whether a pixel belongs to a medicine line. The segmentation model is trained with the above data using a semantic head with two output channels. The predicted medicine label masks obtained from this model may not always respect text boundaries, and hence we use a generic text detector in the OCR pipeline to detect text and refine the boundaries. Then, we crop out the detected bounding box from the original image $\boldsymbol{x}$ and send only those lines to the line recognizer.  As these lines correspond to a special domain of medicine names, we can inject that knowledge to the OCR using a LM.

\subsection{Domain-specific Language Model} \label{sec:med_lm}
In OCR decoder, we can incorporate a LM to correct some of the OCR errors. Specifically, the decoded string $Y^*$ can be obtained as follows:
\begin{equation}
    Y^* = \argmax_Y P(Y|X) P(Y)^\alpha 
    \label{eqn:lm_decoder}
\end{equation}
where $P(Y)$ is obtained from the LM denoting the probability of occurrence of a certain string $Y$ in the dataset, $\alpha$ is the weight applied on the LM, and $X$ is the input. In a generic OCR model, $P(Y)$ is trained on a large corpus of text such that it represents a diverse set of documents. Particular to our use case, once we have detected the medicine lines as discussed in the previous section, we need only medicine line specific knowledge while decoding the OCR output. However, medicine line patterns occurring in handwritten prescriptions often do not appear in normal text. It is also difficult and expensive to acquire and annotate such large corpora of handwritten prescriptions from which we can learn medicine line specific LMs. We inject domain knowledge to solve this problem. 

In order to gather medicine line specific text data, we defined a probabilistic program from which we can sample data and learn a character based LM. Medicine lines written by doctors often have a few elements - a enumeration token (-, ., numbers, etc), followed by the type of medicines (injection, tablet, etc.), the root name of the medicine, and then the suffix. These altogether comprise a single medicine name line. Note that some of these entities other than the root word may not appear in all prescriptions. With this domain knowledge, we can define a probabilistic program as shown in top-right portion of Figure \ref{fig:framework}. The program starts from the START node and ends at the END node, and concatenates the output of each node with spaces in between. To sample a medicine name line, the program takes as input the medicine name and the type of the medicine, both of which appears in the vocabulary of medicines. We can create an exhaustive set of all possible medicine name lines, and then train a character based n-gram LM on that text corpus. Note that as we do not have the exact probabilities of the different transitions, we use equi-probable transitions between nodes, as well as for any choices in the nodes.

In OCR, as decoding is done at a character level, we need character LMs, unlike recent advanced large LMs which operate on word or sub-word tokens. There are also character LM using transformers, but those are generally useful for longer context. But, in our case, medicine names on average are only 7 characters long. Moreover, using such a large model takes a lot more inference time. Hence we stick to an n-gram model.

\subsection{In-Vocabulary Prediction}
In many entity extraction tasks, such as medicine name prediction studied in this paper, the entities often belong to either from a fixed vocabulary, or are defined by a regular expression. However, the OCR predictions will not be constrained to our medicine vocabulary. To constrain that, we can make a nearest neighbor edit distance search for each medicine line text and the medicine vocabulary. However, as we discussed before, it would not respect the model's confidence. Thus, we use the top-k path decoding as a robust method. Specifically, for each line, we decode the top-k predictions, and then find all the text which have an exact match with one of the medicine names from the vocabulary. Then, we take a majority voting of all these matched names, and that becomes the prediction for every line. It is possible that for some of the detected medicine lines, we do not find any match for any of the top-k prediction. These detected medicine lines would not have any output prediction. We find this method to be more effective compared to edit distance based matching with the top-1 prediction, or predicting only the first match from the top-k predictions, as shown in Section \ref{sec:exp}.

\section{Experiments} \label{sec:exp}

We first introduce the dataset and implementation details, before sharing the experimental results and rigorous ablations to understand the efficacy of the framework. 

\subsubsection{Prescription Image Dataset:} 
\begin{figure*}[t]
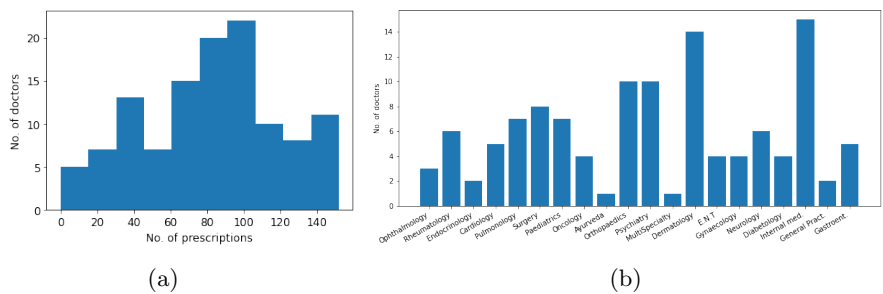

    \small
     \centering
     \begin{subfigure}[t]{0.35\textwidth}
        \centering
        \includegraphics[height=3.25cm]{figures/prescriptions_vs_doctors.pdf}
        \caption{}
        \label{fig:prescriptions_vs_doctors}
     \end{subfigure}
     \begin{subfigure}[t]{0.64\textwidth}
         \centering
         \includegraphics[height=3.25cm]{figures/doc_specialties.pdf}
         \caption{}
         \label{fig:doc_specialties}
     \end{subfigure}
     \caption{\small (a) This plot shows the number of prescriptions per doctor in the dataset, (b) This plot shows the number of doctors per specialty. }
    \label{fig:doctor_details}
\end{figure*}

We use a dataset of handwritten prescriptions to validate the methodology outlined and evaluate the performance of the models. A few example images from the dataset are shown in Figure \ref{fig:sample_images}. The dataset contains 9645 images written by 117 doctors. Table \ref{table:dataset_deets} outlines some of the details of the dataset, and Figure \ref{fig:prescriptions_vs_doctors} shows the distribution of prescription images per doctor. We use 80\% of the dataset to train our models, and 20\% for evaluation. There is no overlap between the doctors between the training and the test set at each iteration, ensuring that our results capture understanding across different handwriting styles. Each image in the dataset has a list of medicine names appearing in them, which we call weak labels, without any positional information. However, just for evaluation, we strongly annotate $500$ images from the evaluation set to evaluate the segmentation performance. Prescriptions generally have multiple other sections as well (although unstructured in free-form), and Table \ref{table:coverage} shows the percentage of images which have other sections such as lab/scans reported, observations and vitals. 
Also, note that any and all personally identifiable information was removed from the data prior to it being provided to the authors for this study. 

\begin{table}[t]
    \scriptsize
	\caption{\small (a) Statistics of the prescription dataset. (b) Coverage of different sections in prescriptions. }
    \begin{subtable}{.5\linewidth}
			\centering
			\caption{}
			\renewcommand{\arraystretch}{1.2}
	        \setlength{\tabcolsep}{4pt}
			\begin{tabular}{l|c}
		    \toprule
		  \# Images & 9645 \\
		  \midrule
		  \# Doctors & 117 \\
		  \midrule
		  Avg. medicines / image & 4.5 \\
		  \midrule
		  Avg. images / doctor & 82.4\\
		   \bottomrule
	    \end{tabular}
	    \label{table:dataset_deets}
    \end{subtable}%
    \begin{subtable}{.5\linewidth}
			\centering
			\caption{}
			\renewcommand{\arraystretch}{1.2}
	        \setlength{\tabcolsep}{4pt}
			\begin{tabular}{l|c}
		    \toprule
		  Lab/Scan & 70.4\% \\
		  \midrule
		  Medicine & 100\% \\
		  \midrule
		  Observation & 99.9\% \\
		  \midrule
		  Vital & 40.5\%\\
		   \bottomrule
	    \end{tabular}
	    \label{table:coverage}
    \end{subtable}%
\end{table}

\subsubsection{Medicine Vocabulary:} We also use a medicine name vocabulary consisting of more than 90,000 medicine names. We use this to generate synthetic medicine name lines and train the character based medicine LM. This vocabulary is also used to make the in-vocabulary predictions.

\subsubsection{Evaluation Protocol:} We evaluate all models on test set of the dataset mentioned above. To evaluate the performance of the segmentation model, we use mean intersection over union (mIoU) as used in the segmentation literature \cite{chen2017deeplab}. To evaluate the performance of the end-to-end medicine name prediction model, we use the mean jaccard index, over all the images. We also use two other metrics namely mean precision and mean recall, and the mean jaccard index can be considered as a combination of both these metrics. These are defined as follows - 
\begin{eqnarray}
    \text{Mean Jaccard Index (mJI)} = \frac{1}{M} \sum_{i=1}^M \frac{|P_i \cap G_i|}{|P_i \cup G_i|} \\
    \text{Mean Precision (mP) } = \frac{1}{M} \sum_{i=1}^M \frac{|P_i \cap G_i|}{|P_i|}, \\
    \text{Mean Recall (mR)} = \frac{1}{M} \sum_{i=1}^M \frac{|P_i \cap G_i|}{|G_i|}
\end{eqnarray}
where $P_i, G_i$ are the predicted and ground truth list of medicines for the $i^{th}$ image. $M$ is the number of evaluation images. The comparison between the prediction and ground-truths are not case-sensitive, as they are medicine names.

\subsection{Results and Ablation Studies} \label{sec:seg}

\subsubsection{Iterative Training Performance:} 
As discussed in Section \ref{sec:method}, our algorithm for converting weak labels (only medicine names) to strong labels (bounding box annotations for each medicine name) involves two labeling functions - OCR and Segmentation Labeling Function, where the latter can be applied iteratively. The number of images auto-labeled by the labeling functions increases with iterations, and hence the performance of both the medicine line segmentation model as well as the medicine name prediction model increases with subsequent iterations. We highlight this in Table \ref{table:seg_iter}. Iteration 1 shows the performance on only OCR Labeling Function, and Iteration $\geq 2$ shows the performance on multiple iterations of Segmentation Labeling Function. For a significant number of prescriptions, it is difficult to decipher some of the medicine names, even when we use a high value of top-k (k=20,000 in our experiments) decoded outputs per line. For Iteration 1, the number of auto-labeled prescriptions is $<25\%$ of the training set. This shows the difficulty level of the problem at hand. Note that the train sets are used to train only the medicine line segmentation model and not the lines recognizer of the OCR, thus it can be with any off-the-shelf OCR model. 

The segmentation performance as well as the medicine name performance improve over iterations but saturates from Iteration 3. Note that mIoU computes the performance for every pixel, but normally a small change in the final bounding box do not have a lot of impact on the medicine name prediction, as long as they encapsulate the text within it. We also show the upper bound performance of medicine line recognition by using ground-truth medicine bounding boxes only while evaluating. As we can see, our algorithm with just using weak labels can reach within a few points of the strongest upper-bound with strong labels.
\begin{table}[t]
            \small
			\caption{\small Performance over iterations of the proposed framework. Iter 1 represents learning from only the OCR Labeling Function and iter $\geq 2$ shows the performance after iteratively including the Segmentation Labeling Function. The medicine name performances are only for topk=1. GT bbox shows the performance when the groundtruth bounding boxes are provided for medicine names only during evaluation.}
			\label{table:seg_iter}
			\centering
			\renewcommand{\arraystretch}{1.2}
	        \setlength{\tabcolsep}{4pt}
			\begin{tabular}{l|cccc}
		    \toprule
		    Iteration & 1 & 2 & 3 & GT bbox\\
		    \midrule
		    Train data (\%) & 24.4 & 66.3 & 90.2 & - \\
		    \midrule
		    Segmentation  (mIoU) & 72.6 & 77.9 & 77.2 & 100\% \\
		    Medicine Name (mJI) & 44.8 & 45.9 & 45.9 & 49.8\% \\
		   \bottomrule
	    \end{tabular}
\end{table}

\subsubsection{Cues for medicine name segmentation:}
\begin{figure*}[t]
     \centering
     \begin{subfigure}[t]{0.32\textwidth}
        \centering
        \includegraphics[scale=0.13]{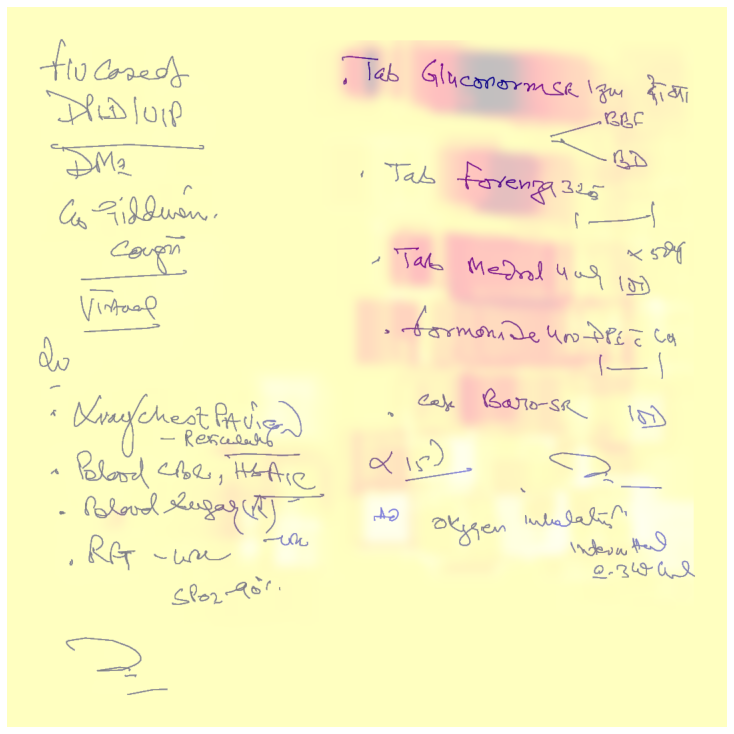}
        \caption{}
        \label{fig:cue1}
     \end{subfigure}
     \begin{subfigure}[t]{0.32\textwidth}
         \centering
         \includegraphics[scale=0.13]{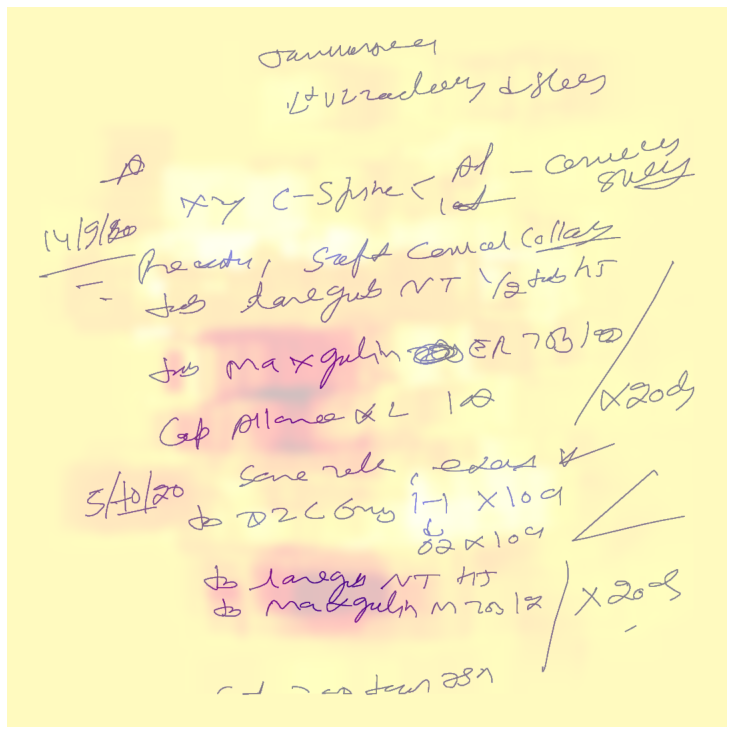}
         \caption{}
         \label{fig:cue2}
     \end{subfigure}
     \begin{subfigure}[t]{0.32\textwidth}
         \centering
         \includegraphics[scale=0.13]{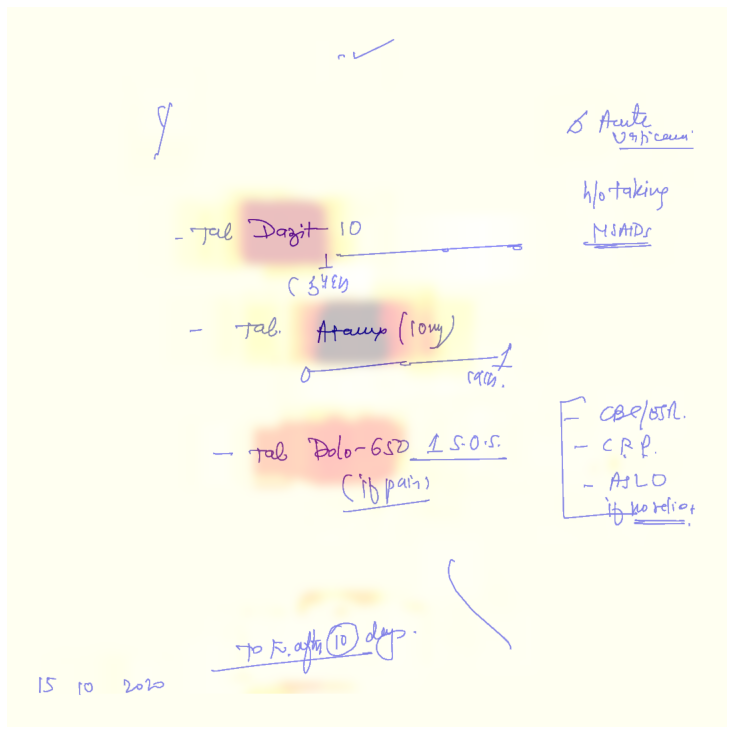}
         \caption{}
         \label{fig:cue3}
     \end{subfigure}
     \vspace{-2mm}
     \caption{\small Cues needed by the segmentation network. Deeper color denotes lower performance when a patch around that is removed. A few parts of the image other than the medicine names, such as hyphens, Tab, Cap, etc., also appear to be darker, which are some of the cues that the model looks at to determine whether it is a medicine line.}
    \label{fig:cue}
\end{figure*}

Unlike a generic text detector, specifically detecting medicine lines can be challenging, as handwritten prescriptions do not have any specific structure or location in the page. However, the segmentation model is still able to predict the location of the medicine lines with high performance as shown in Table \ref{table:seg_iter}. In order to understand the cues the segmentation model uses to segment the medicine names, we do the following experiment. Given a test image $\boldsymbol{x}$, using a sliding window, we remove square patches from the image to remove potential cues, one at a time. Consider $\boldsymbol{x}_{i,j}$ as the image when patch at location ($i,j$) is removed. We can run the segmentation model on this image, $\mathcal{M}(\boldsymbol{x}_{i,j})$ and obtain the mIoU. For every location ($i,j$), on the image, we can obtain the model's performance drop when a patch around that is removed, and then display that as a heatmap. A drop in performance in certain regions of this image depicts the regions necessary for the segmentation model to segment the medicine names correctly. As we can see in Fig. \ref{fig:cue}, the model is clearly utilizing cues from visual features surrounding medicine lines such as starting of a line like Tab, Cap, hyphens, etc. These observations are aligned with what a pharmacist or even non-domain experts look to determine medicine lines, as in most cases the handwriting is illegible. These key demarcations serve as strong signals to recognize medicine lines, after which we can condition our knowledge to medicine names to enhance line recognition.

\subsubsection{Contribution of medicine LM and segmentation model:} 
\begin{figure*}[t]
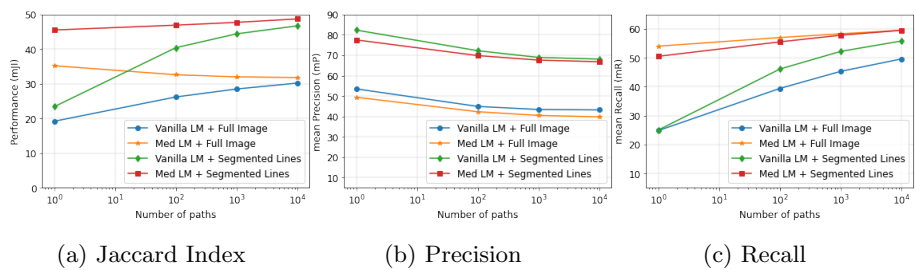

     \centering
     \begin{subfigure}[t]{0.32\textwidth}
        \centering
        \includegraphics[scale=0.30]{figures/num_paths_ablation_iou.pdf}
        \caption{Jaccard Index}
        \label{fig:num_paths_iou}
     \end{subfigure}
     \begin{subfigure}[t]{0.32\textwidth}
         \centering
         \includegraphics[scale=0.30]{figures/num_paths_ablation_precision.pdf}
         \caption{Precision}
         \label{fig:num_paths_precision}
     \end{subfigure}
     \begin{subfigure}[t]{0.32\textwidth}
         \centering
         \includegraphics[scale=0.30]{figures/num_paths_ablation_recall.pdf}
         \caption{Recall}
         \label{fig:num_paths_recall}
     \end{subfigure}
     \vspace{-2mm}
     \caption{\small Jaccard index, precision and recall comparison using different language models and inputs (medicine line segmented and full page). The medicine LM on segmented medicine lines works the best, the top-1 of which is better than the top-1000 of the vanilla LM. Applying the medicine LM on the entire image decreases the precision of the predictions, as it hallucinates medicine names in the rest of the prescription. }
    \label{fig:num_paths_ablation}
\end{figure*}

Here we show how selectively injecting medicine LMs can offer a significant improvment in performance. The vanilla LM is trained on a generic corpus of text from the Latin script. However, the medicine name LM is trained as discussed in Section \ref{sec:med_lm}. The performance improves with path length for both the models but for the medicine LM, the top-1 path itself performs much better than top-1000 path for the vanilla LM (Fig.~\ref{fig:num_paths_ablation}). This also reduces the compute time in decoding the top-k paths from the logits, which is linear in the number of paths.

Moreover, segmenting and selectively injecting the LM plays a critical role on the performance, and MedLM + Segmented Lines perform the best. Applying the MedLM on the full image actually reduces the precision significantly, but improves the recall slightly as expected, but reducing the overall metric, i.e., jaccard index. This shows that selectively injecting the LM is important, otherwise it can mess up the rest of the prescription, and hallucinate medicine names from them. 

\begin{figure*}[t]
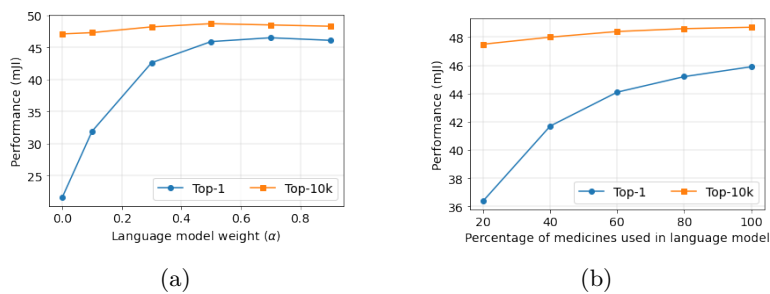

     \centering
     \begin{subfigure}[t]{0.45\textwidth}
        \centering
        \includegraphics[scale=0.33]{figures/lm_ablation_.pdf}
        \caption{}
        \label{fig:lm_weight}
     \end{subfigure}
     \begin{subfigure}[t]{0.45\textwidth}
         \centering
         \includegraphics[scale=0.33]{figures/med_percentage_ablation.pdf}
         \caption{}
         \label{fig:med_percentage}
     \end{subfigure}
     \vspace{-3mm}
     \caption{\small (a) Ablation of performance with weight on the language model $\alpha$. $\alpha=0$ denote the performance of only the optical model. (b) Ablation of fraction of medicine names used to train the medicine language model. We present the performance when top-1 and top-10k paths are used to predict after vocabulary matching.}
    \label{fig:lm_ablation}
\end{figure*}

\subsubsection{Performance with varying weight on LM:} 
The weight $\alpha$ in Eqn \ref{eqn:lm_decoder} on the LM scores can have an impact on the final performance. A low weight may lead to no improvement beyond the optical model's prediction, and a high weight may not ground the output to the actual text on the image. Figure \ref{fig:lm_weight} shows an ablation of the medicine name prediction performance on the LM weight. Note that the changes in performance is much lower for top-10k paths than for top-1 path, as only the first path in the top-10k path is affected by the LM because for paths $>1$, the predictions come from the top-k decoded paths which is based on only the logits without any LM scoring. Nonetheless, we see that the performance of both the models are very close after a certain value of $\alpha$. 

\subsubsection{Varying the vocabulary of the LM:} 
The medicine names used in generating the synthetic lines can have an impact on the quality of the medicine name LM. Here we also show how the performance varies as we increase the number of medicine names used to train the medicine LM. Figure \ref{fig:med_percentage} presents the results for top-1 and top-10k with different size of medicine name dataset. The performance improves as we add more medicines, but starts saturating after a certain point.

\subsubsection{Performance with different n-gram models:} 
The n-gram LM involves a parameter $n$, which is the number of history characters the model looks to obtain the score of the next character. We created multiple n-gram models on the synthetically generated medicine line text data, and show the results in Table \ref{table:ngram}. More context definitely helps in performance, but it saturates after $n=7$. This is also intuitive as the length of the medicine names is around $7.9$ on average. 

\begin{table}[t]
			\caption{\small Ablation of different n-gram models trained on medicine line data.}
			\label{table:ngram}
			\centering
			\renewcommand{\arraystretch}{1.2}
	        \setlength{\tabcolsep}{4pt}
			\begin{tabular}{l|ccccccc}
		    \toprule
		     & n=3 & n=5 & n=7 & n=9 \\
		    \midrule
		    Top-1 (mJI) (\%) & 27.2 & 41.5 & 45.9 & 45.9 \\
		    Top-10k (mJI) (\%) & 47.4 & 48.1 & 48.7 & 48.7 \\
		   \bottomrule
	    \end{tabular}
\end{table}

\subsubsection{Predicting In-Vocabulary Words:} \label{sec:med_matching}
In the final step of our algorithm to predict medicine names, we only predict those words where we find a direct match with one of the elements of the medicine vocabulary. As discussed before, finding a match for only the top-1 prediction may not be the best. Thus, we decode until top-k and find matches for all the text. As the top-k decoding is directly dependent on the output of the model, such a matching respects the model's predictions. We then take a majority voting of all the matches and that becomes the final predicted medicine for a line. Note that some lines may not have any prediction at all. In this section, we compare multiple strategies of predicting in-vocabulary words in Table \ref{table:invocab_pred}. Top-1 represents an exact match with the first path, top-1 edit distance finds the nearest prediction from the vocabulary by edit distance, top-k denotes we decode the top-k outputs but stop when we find the first exact match, and finally top-k+majority is the algorithm we use, where we decode all the top-k lines and take a majority voting of all the exact matches.

Note that top-1-edit has the same jaccard index as top1, but the former has lower precision with higher recall than top-1, as expected, because it predicts beyond exact matches. We tried with multiple thresholds for edit distance, and found that 85\% normalized distance performs the best. Increasing the threshold, i.e., allowing more matches significantly reduces the precision, at the gain of the recall, but hurting the overall performance. This is because of the intuition we discussed earlier that topk decodings respect the model's confidence, but edit distance treats every replacement with the same cost. 

\begin{table}[t]
			\caption{\small Ablation of different algorithms to predict medicine names. We use $k=1e4$.}
			\vspace{2mm}
			\label{table:invocab_pred}
			\centering
			\renewcommand{\arraystretch}{1.2}
	        \setlength{\tabcolsep}{4pt}
			\begin{tabular}{lcccc}
		    \toprule
		    & top-1 & top-1-edit & top-k & top-k+majority \\
		    \midrule
		    Jaccard Index & 45.9 & 45.8 & 46.9 & 48.7\\
		    Precision & 76.9 & 68.6 & 64.8 & 66.8\\
		    Recall & 51.0 & 54.4 & 58.5 & 59.5 \\
		   \bottomrule
	    \end{tabular}
\end{table}

\subsection{Error-mode analysis}
The two types of errors possible are - medicine names predicted but not in the ground-truth (type I) and medicine names in the ground-truth but not predicted (type II). In our framework, there are two reasons behind the errors - segmentation network and OCR. If a medicine name is not segmented, then it leads to a type-II error. OCR errors contributes to the rest (type I and type II), a majority of which is contributed by misinterpreting very similar looking medicines such as emtel vs entel, eenosol vs eenasof, folvite vs folite, paro vs baro, zincovit vs zincort, aloliv vs alcoliv. Also we observe that the doctor can commit spelling mistakes, or vaguely write a medicine name, where only the first few characters are recognizable. To correct such errors, pharmacists generally use other contexts such as observation. Learning such contexts would need a lot more data, and injecting higher-level domain knowledge.

\section{Conclusion}

In this paper, we looked into the problem of extracting medicine names from inscrutable handwritten prescriptions. Our algorithm can selectively infuse domain knowledge to specific portions of a document to significantly improve the performance. We developed a framework that can learn to detect regions of interest from just weak labels, and also learn a medicine language model using synthetically generated text lines using probabilistic programs. The idea is generic enough to be applied to a variety of other types of documents, such as handwritten forms.

\subsubsection{Acknowledgement:}
We thank Srujana Merugu, Ansh Khurana, Manish Gupta, Harsh Dhand and Shruti Garg for all the support and discussions during the course of this project. Without their effort, this project would not have been possible.



{\small
\bibliographystyle{splncs04}
\bibliography{egbib}
}





\end{document}